\documentclass[conference]{IEEEtran}
\IEEEoverridecommandlockouts
\usepackage{cite}
\usepackage{amsmath,amssymb,amsfonts}
\usepackage{algorithmic}
\usepackage{graphicx}
\usepackage{textcomp}
\usepackage{xcolor}
\def\BibTeX{{\rm B\kern-.05em{\sc i\kern-.025em b}\kern-.08em
    T\kern-.1667em\lower.7ex\hbox{E}\kern-.125emX}}
\begin{document}

\title{Towards Structured Evaluation of Deep Neural Network Supervisors}

\author{
\IEEEauthorblockN{Jens Henriksson\IEEEauthorrefmark{1}, Christian Berger\IEEEauthorrefmark{2}, Markus Borg\IEEEauthorrefmark{3}, Lars Tornberg\IEEEauthorrefmark{4},\\ Cristofer Englund\IEEEauthorrefmark{3}, Sankar Raman Sathyamoorthy\IEEEauthorrefmark{5}, Stig Ursing\IEEEauthorrefmark{1}}\\
\IEEEauthorblockA{\IEEEauthorrefmark{1}Semcon AB, Gothenburg, Sweden, Email: \{jens.henriksson, stig.ursing\}@semcon.com}
\IEEEauthorblockA{\IEEEauthorrefmark{2}University of Gothenburg and Chalmers Institute of Technology, Sweden, Email: christian.berger@gu.se}
\IEEEauthorblockA{\IEEEauthorrefmark{3}RISE Research Institutes of Sweden AB, Lund and Gothenburg, Sweden, \{markus, cristofer\}@ri.se}
\IEEEauthorblockA{\IEEEauthorrefmark{4}Machine Learning and AI Center of Excellence, Volvo Cars, Gothenburg, Sweden, Email: lars.tornberg@volvocars.com}
\IEEEauthorblockA{\IEEEauthorrefmark{5}QRTech AB, Gothenburg, Sweden, Email: sankar.sathyamoorthy@qrtech.se}}

\maketitle

\begin{abstract}
Deep Neural Networks (DNN) have improved the quality of several non-safety related products in the past years. However, before DNNs should be deployed to safety-critical applications, their robustness needs to be systematically analyzed. A common challenge for DNNs occurs when input is dissimilar to the training set, which might lead to high confidence predictions despite proper knowledge of the input. 

Several previous studies have proposed to complement DNNs with a supervisor that detects when inputs are outside the scope of the network. Most of these supervisors, however, are developed and tested for a selected scenario using a specific performance metric. In this work, we emphasize the need to assess and compare the performance of supervisors in a structured way. We present a framework constituted by four datasets organized in six test cases combined with seven evaluation metrics.  

The test cases provide varying complexity and include data from publicly available sources as well as a novel dataset consisting of images from simulated driving scenarios. The latter we plan to make publicly available. Our framework can be used to support DNN supervisor evaluation, which in turn could be used to motive development, validation, and deployment of DNNs in safety-critical applications.
\end{abstract}


\section{Introduction}

Deep Learning (DL) has produced groundbreaking results in recent years. Building on the increasing availability of data and massive parallel processing power, DL has revolutionized research topics such as computer vision~\cite{he_deep_2016}, machine translation~\cite{cho2014learning}, and voice recognition~\cite{hinton2012deep}. However, while DL can achieve accuracy beyond human capabilities for specific tasks~\cite{he_delving_2015}, it is not yet clear how development organizations should systematically approach testing of DL-based systems.

Since DL testing in a structured way is an open challenge similar to the broader concept of Machine Learning (ML) Verification \& Validation (V\&V)~\cite{borg_safely_2019}, DL-enabled systems are currently mostly deployed in non-critical application domains. Using DL to place online advertisements to maximize conversion rates in e-commerce introduces limited risks, similar to using DL in smart-phone augmented reality apps to amuse users. But what about safety-critical applications for which human health, the environment, or large financial assets are at stake? 
%In this paper, we focus on such a safety-critical DL application: Vehicle  surroundings perception by using DL to obtain awareness of actors in traffic.
%\lars{is this really the focus in this paper or should we just state that as one of the use cases?} \jens{agree with Lars}\cbe{we should name it as one possible use and demonstration case}\lars{I moved this to the scope section. Is that ok?}

\subsection{Problem Domain and Motivation}
The limited transparency of DL is one of the major impediments of using DL in domains that require a systematic safety certification. From the perspective of the safety standard used for automotive software (ISO~26262\footnote{Note that the current paper was written before ISO/PAS 21448 Road vehicles -- Safety of the intended functionality was published in January 2019. Any future work on the topic of ML and automotive safety should consider the new standard as a point of reference.}), using DL introduces a major paradigm shift compared to conventional software systems~\cite{falcini_deep_2017,salay_analysis_2017,henriksson_automotive_2018}. Safety certification requires a convincing argument, which is structured in a safety case to outline why a system is safe~\cite{kelly1999arguing}. Instead of human-readable source code, state-of-the-art DL might be composed of hundreds of millions of unexplainable parameter weights, necessitating novel approaches for V\&V to evolve for DL~\cite{borg_safely_2019}. Thus, several techniques mandated by ISO~26262 cannot be directly applied to DL, e.g., source code reviews and exhaustive coverage testing, which aim at increasing function readability or maintainability that is less relevant for Neural Networks (NN) in general and Deep Neural Networks (DNN) in particular. To achieve convincing argumentation of safety for DL enabled systems, we believe some form of run-time NN supervision is required. 

% \jens{removed final part. Still believe we need to have a final touch on this paragraph, that highlight the importance of DL supervisor} % as a full coverage of a neural network (NN) can be achieved easily, which is, though, meaningless. \ce{Should we explain why it is meaningless??}. Hence, for safety-critical applications, the use of supervisors for DL is gaining interest to enable deployment in critical contexts.

We envision a DL supervisor that continuously performs novelty detection~\cite{pimentel_review_2014} during run-time to detect when input does not resemble the training data. Upon such a detection the supervisor should re-direct the software execution to a \textit{safe-track}, which should be implemented without ML to offer a graceful degradation. This is a common safety engineering concept~\cite{ploeg2015graceful} that has been proposed also for NN-based control system~\cite{EnglundComb2007}. Moreover, the safe track could be targeted by conventional approaches to safety-critical software engineering and can thus obtain safety certification.

\subsection{Research Goal and Research Questions}
There is currently no established framework for how to systematically evaluate DL supervisors. Hence, we cannot draw firm conclusions on which DL supervisor is most appropriate. Even worse, without a structured approach for testing, we cannot conclude whether any DL supervisors yield promising results at all. As such, the goal of this work is to develop such a framework. %\cbe{added this sentence to formulate clearly a research goal (that was not explicitly stated before); ALL: refine if needed.} %Good! 
Based on these observations, we formulate the following research question to guide our work. 
\begin{itemize}
    \item RQ: How could the evaluation of different DL supervisors be supported?
\end{itemize}

\subsection{Contributions}
We present an evaluation framework for DL supervisors. Our approach has three fundamental components: (i) definition of metrics, and (ii) test case setups; in addition, (iii) we show how to use the evaluation method on two test cases by developing two supervisors. The first supervisor shows how to only utilize the softmax output of a Convolutional Neural Network (CNN) classifier~\cite{DBLP:journals/corr/SimonyanZ14a} for anomaly detection, whereas the second one uses a variational autoencoder~\cite{DBLP:journals/corr/KingmaW13} for the same purpose. 

%\jens{fragment for conclusion: ` is an initial version of , i.e., safety cages around machine learning classifiers'}

%\textit{cagedsMiLe}\mb{If we want a name...}\cbe{are there more catchy names as we want to take the world by storm? :-)} -- a framework for evaluating DL supervisors to answer our research question. Our approach has two fundamental components: 1) metrics, and 2) evaluation setup. \mb{This paragraph must be written when there are results in the paper.}

%\jens{I suggest that this snippet goes to `Threats to Validity': In line with common practice in evidence-based medicine~\cite{hoffmann_better_2014}, we report the evaluation framework prior to conducting the actual evaluative study. This practice enables peer-review of its design, prior to running effort-intensive studies with large amounts of data annotated by humans.}

\subsection{Scope and Limitations}
In our work, we consider primarily two example supervisors for illustrations: A softmax threshold technique and a variational autoencoder. These examples demonstrate certain aspects of DL for (i) non-safety-critical applications, and (ii) examples from a self-driving vehicular application where the supervisor is used to obtain awareness of novel traffic scenarios. While we demonstrate different supervisors applied on different test cases to show transferability of our ideas, we are not yet claiming generalizability towards having another instrument ready to be used within safety case argumentation in the context of ISO~26262 for example.

\subsection{Structure of the Article}
The rest of this paper is structured as follows: Section~\ref{sec:rw} presents related work on DL testing and DL supervisors. Section~\ref{sec:eval} describes our approach to structured evaluation, Section~\ref{sec:examples} presents two example applications of the evaluation method. In Section~\ref{sec:disc}, we discuss the practical implications of our work, both for research and practice. Finally, Section~\ref{sec:conc} concludes our work and outlines directions for future work. %TODO: Check again

\section{Related Work}
\label{sec:rw}

The area of testing DL-based approaches and NN has gained a lot of attention recently to improve the way to evaluate performance and robustness. One goal of the testing is to allow DL-based approaches in areas that are typically very sensitive to changes in the input data or where unexpected behaviour or wrong decisions can have severe consequences. Recent studies like DeepXplore from Fei et al.~\cite{deepXplore} or its extension DLFuzz from Guo et al.~\cite{DLFuzz} aim at measuring and improving neuron coverage. Such approaches look at a resulting NN in general to find or generate test data for cases that have not been covered from the test dataset; reported improvements are in the range of around 2-3\%. While these approaches look at the NN itself, our goal is to detect unwanted behaviour based on the input and an NN's output data to enable the design of appropriate supervisors. Kim et al.~\cite{GuidedDLTesting} propose an approach that evaluates the way a given input from the training data is influencing the behaviour of the NN, to guide testers to select test data resulting in a certain \emph{surprise} in an NN as an adequacy criterion. %\lars{this sentence needs to be broken down I think.}

%\subsection{Testing methods}
Ma et al.~\cite{ma2018deepgauge} propose DeepGauge, a set of multi-granularity testing criteria for DL systems, which aims at rendering a multi-faceted portrayal of the testbed. %$https://arxiv.org/pdf/1803.07519.pdf$ 
Experimental results show that DeepGauge may be a useful tool for evaluating testing adequacy of NNs. 

% Kim et. al.~\cite{kim2018guiding} propose a novel test adequacy criterion for testing of DL systems, called Surprise Adequacy for Deep Learning Systems (SADL) $https://arxiv.org/pdf/1808.08444.pdf$

Another approach to achieve robustness of an NN-based system is to monitor the performance online. For example, Hendrycks and Gimpel~\cite{hendrycks2016baseline} presented a method looking at the softmax layer of the NN as a prediction probability. They tested with several datasets and show promising results with respect to predicting whether the trained classifier will correctly classify a test example or not, and also to determine whether the test example is in- or out-of-distribution of the training data. % $https://arxiv.org/pdf/1610.02136.pdf$

Liang et al.~\cite{Liang18} propose ODIN, an out-of-distribution detector for neural networks that does not require any change to a pre-trained model. The method is based on the observation that using temperature scaling and adding small perturbations to the input can separate the softmax score distributions of in- and out-of-distribution samples, allowing for more effective detection of unwanted behaviour. % $https://www.researchgate.net/publication/317418851_Principled_Detection_of_Out-of-Distribution_Examples_in_Neural_Networks$

Englund and Verikas~\cite{EnglundComb2007} propose to monitor the error between the predicted output and the measured output of an industrial process to switch between a NN-based controller and a traditional integrating controller. It was experimentally shown that when the direct model had difficulties to predict the process output, also the inverse model had difficulties to predict the control signal and hence, a safe track consisting of an integrating process controller was used to bring the process back to the desired output range. 

Another supervisor approach was proposed by Tranheden and Landgren~\cite{MattiasLudwig18}. The supervisor monitored activations from all layers of the NN and not only the softmax layer as in~\cite{hendrycks2016baseline}. Their conclusion is that the information, which is contained in the different layers of the NN, and how it is represented therein, varies with both network and the problem at hand, and therefore NN-specific supervisors may be needed.  % $http://publications.lib.chalmers.se/records/fulltext/255752/255752.pdf$

%Additional good papers for related work 
 %\begin{enumerate}
%    \item Tornberg et. al : Do you got any paper that presents the VAE based approach you looked into?  \lars{I have added two refs and refer to them in the use case}
%    \item SMILE 1 project mapped out the problematic areas, whereas explainability of the trained models were the most important. They continued with a follow-up project looking at, what they defined as, a safety-cage approach. 
%    \item Braiek and Khomh reports the following:   First, we identify and explain challenges that should be addressed when testing ML programs. Next, we report existing solutions found in the literature for testing ML programs. Finally, we identify gaps in the literature related to the testing of ML programs and make recommendations of future research directions for the scientific community. 
%     \item Amodei et. al. discuss one such potential impact: the problem of accidents in machine learning systems, defined as unintended and harmful behaviour that may emerge from poor design of real-world AI systems. $https://arxiv.org/pdf/1606.06565.pdf$
% \end{enumerate}

% \jens{we should keep one joint Related Work section to provide a comprehensive and not scattered overview}
%\subsection{Related work in SMILE 1}
%\subsection{Related papers we found}

%%%%%%%%%%%%%%%%%%%%%%%%%%%%%%%%%%%%%%%%%%%%%%%%%%%%%%%%%%%%%%%%%%%%%%%%%%%%%%%%%%%%%%
\section{Evaluation Method}\label{sec:eval}

In this section we describe our proposed framework for comparing supervisors. To achieve fair comparison between different methods, they must be compared on common grounds. Differences in setups will affect the performance of the supervisors and also render it impossible to benchmark different supervisors with each other. To reduce the impact of these differences we propose a fixed set of datasets, metrics, and requirements for models under supervision. In the rest of the section, we describe all parts starting with the datasets that we used for experimentation.
%in line with recommendations from Hoffmann et al.~\cite{hoffmann_better_2014}, 

\begin{figure}
    \centering
    \includegraphics[width=.99\linewidth]{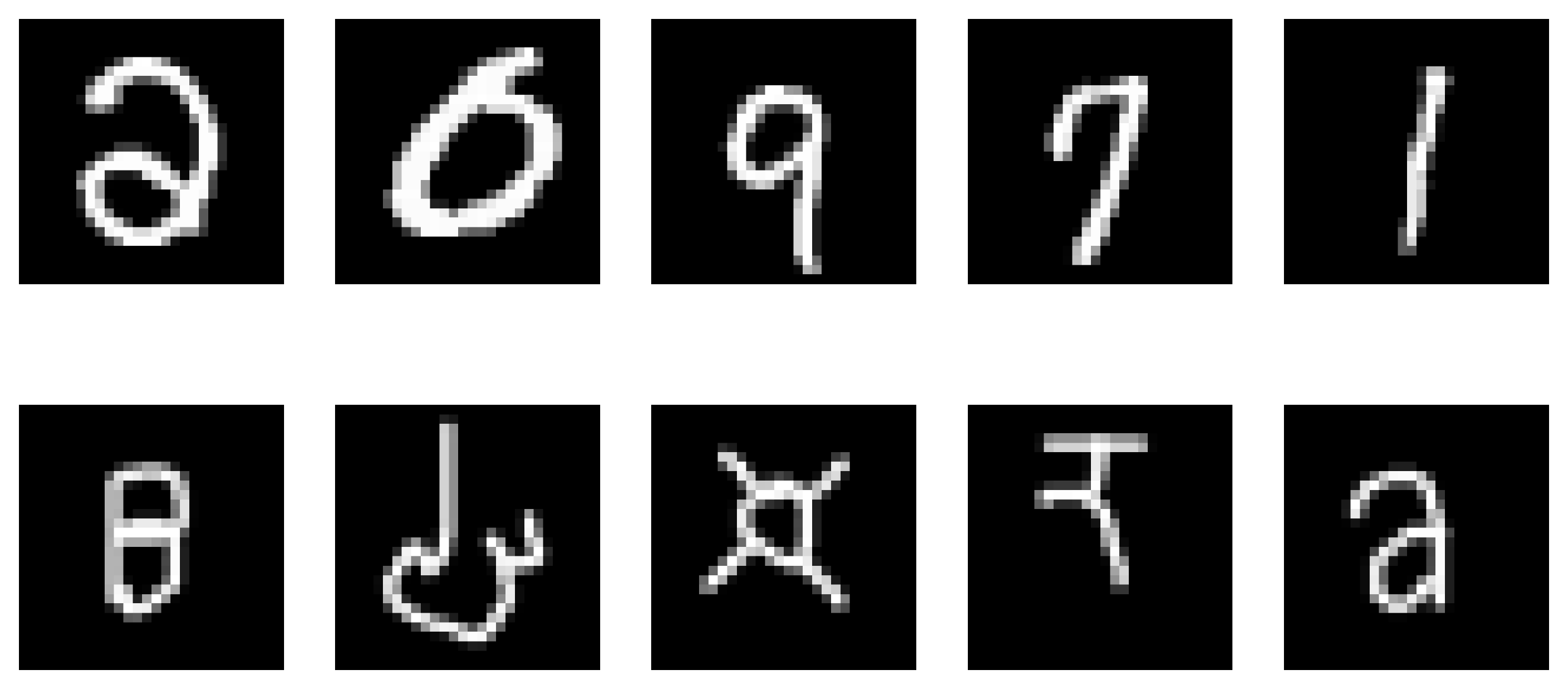}
    \caption{Example images of the training data of MNIST handwritten digits (above) and the outlier samples from Omniglot (below). }
    \label{fig:mnist-omniglot-examples}
\end{figure}

\textbf{Datasets}: Our framework considers four dataset combinations, which together represent six inlier/outlier cases as can be seen in Tab.~\ref{tab:data_sets}. All datasets are partitioned in inliers/outliers, where the training data from the inlier set is used to train the model and the test data from the inlier set in combination with the outlier data is used to evaluate the supervisor performance. The first dataset combination: MNIST/Omniglot aims at introducing a simple case to illustrate the possibility of supervision for simple scenarios. The training data is distributed with train/test setup, and is used as is. For the corresponding outlier data we choose an equal amount of samples as the test data from the Omniglot dataset \cite{Lake1332} which consists of 1,623 handwritten characters from over 50 alphabets. The omniglot images are rescaled to fit the dimension of the MNIST images. Example images of these datasets can be seen in Fig.~\ref{fig:mnist-omniglot-examples}.

\begin{figure}
    \centering
    \includegraphics[width=.99\linewidth]{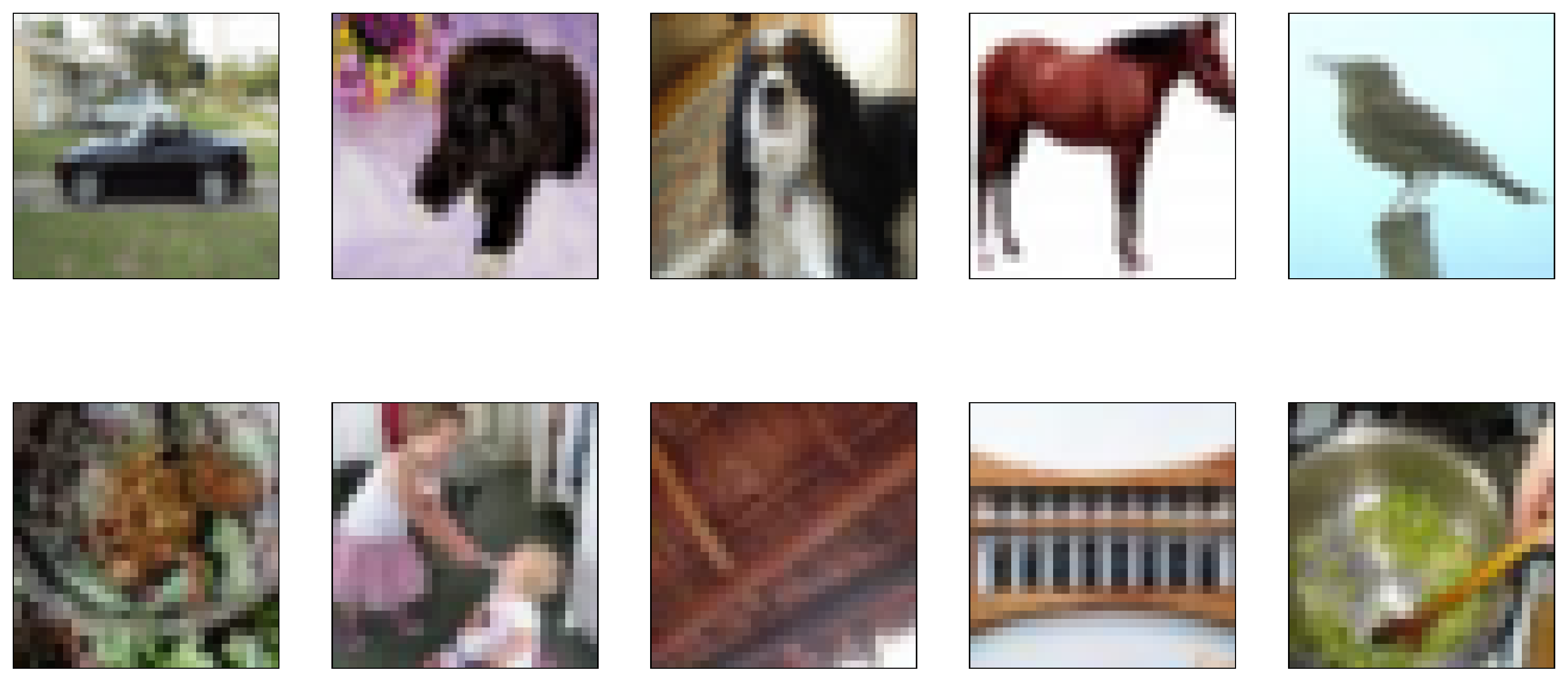}
    \caption{Example images of the training set of CIFAR-10 (above), and outlier samples from the Tiny ImageNet set (below).}
    \label{fig:cifar-tiny-imagenet-examples}
\end{figure}

The second combination is RGB-data from CIFAR-10 \cite{CIFAR10} and Tiny ImageNet (a subset from the ImageNet Large Scale Visual Recognition Challenge \cite{ILSVRC15}). These datasets contain small sized images, where the object in focus is covering the majority of the image as can be seen in Fig.~\ref{fig:cifar-tiny-imagenet-examples} and hence, background information from the images will have less impact on the supervisor. The CIFAR-10 data comes pre-distributed in train and test data, which is also used as is. The outlier samples are selected as the test data from Tiny ImageNet, but resized to match the CIFAR-10 image size. 

The third dataset is created from a simulator as an introduction to more realistic data for self-driving vehicular applications. The dataset consists of realistic driving scenarios that are simulated in the virtual prototyping platform Pro-SiVIC\textsuperscript{TM} from ESI\footnote{https://www.esi-group.com/software-solutions/virtual-environment/virtual-systems-controls/esi-pro-sivictm-3d-simulations-environments-and-sensors}. Pro-SiVIC\textsuperscript{TM} allows us to simulate different environments, weather conditions, vehicle types, and sensor setups. The outputs considered here are from a camera sensor attached to the vehicles. Pro-SiVIC\textsuperscript{TM} also provides pixel by pixel semantic segmentation of the images. For our framework, we have defined the following driving scenarios: As autonomous highway pilots might be one of the first fully self-driving solutions, we have around 7,500 images from driving on a highway during sunny weather as our inlier data. The outlier data consists of scenarios including driving in foggy conditions (around 780 images) and driving in urban areas (around 480 images). Figs.~\ref{fig:procivic_inlier}-\ref{fig:procivic_outlier_urban} show selected samples from the inlier and outlier data.

\begin{figure}
  \centering
  \includegraphics[width=0.32\linewidth]{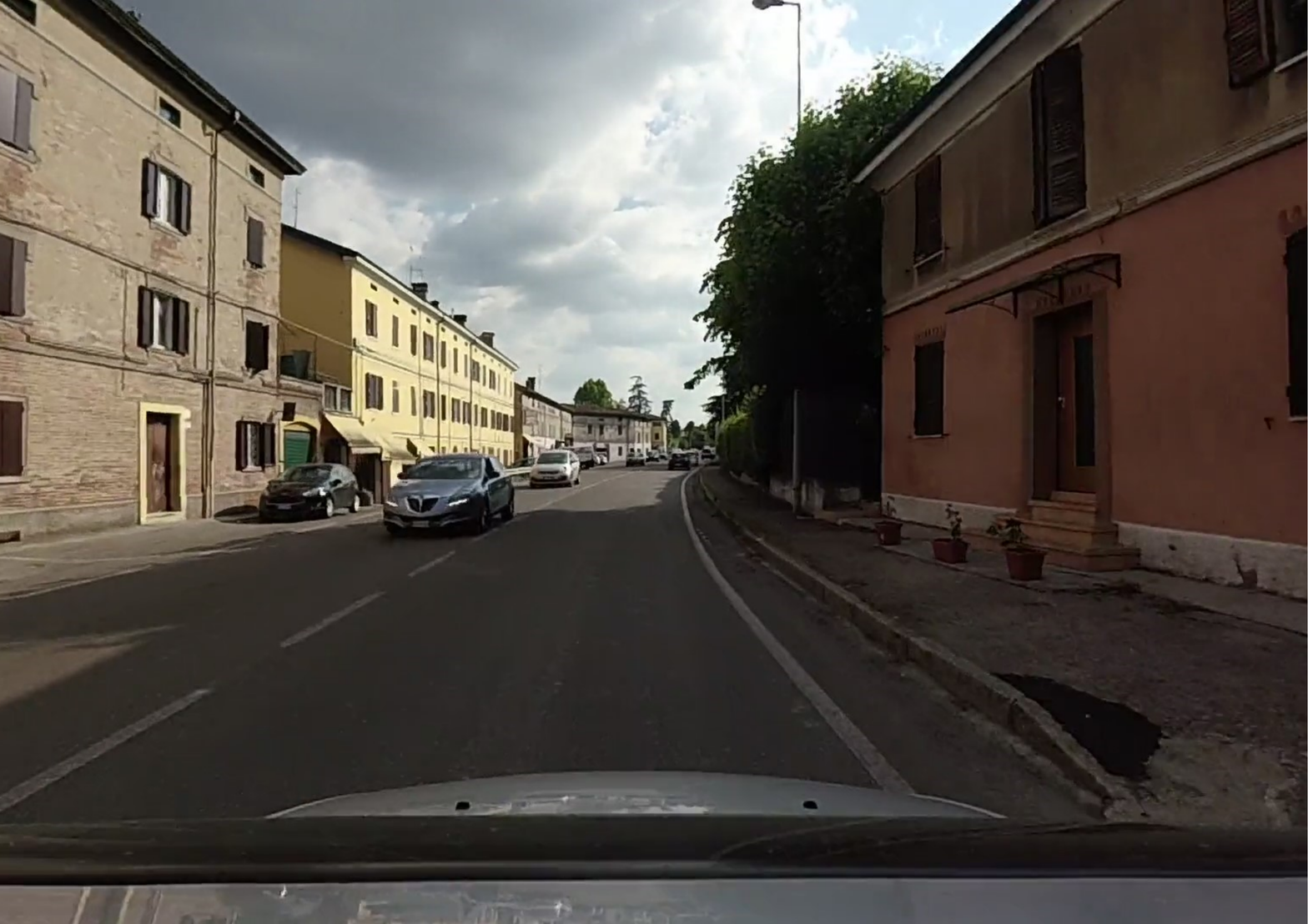}
  \hfill
  \includegraphics[width=0.32\linewidth]{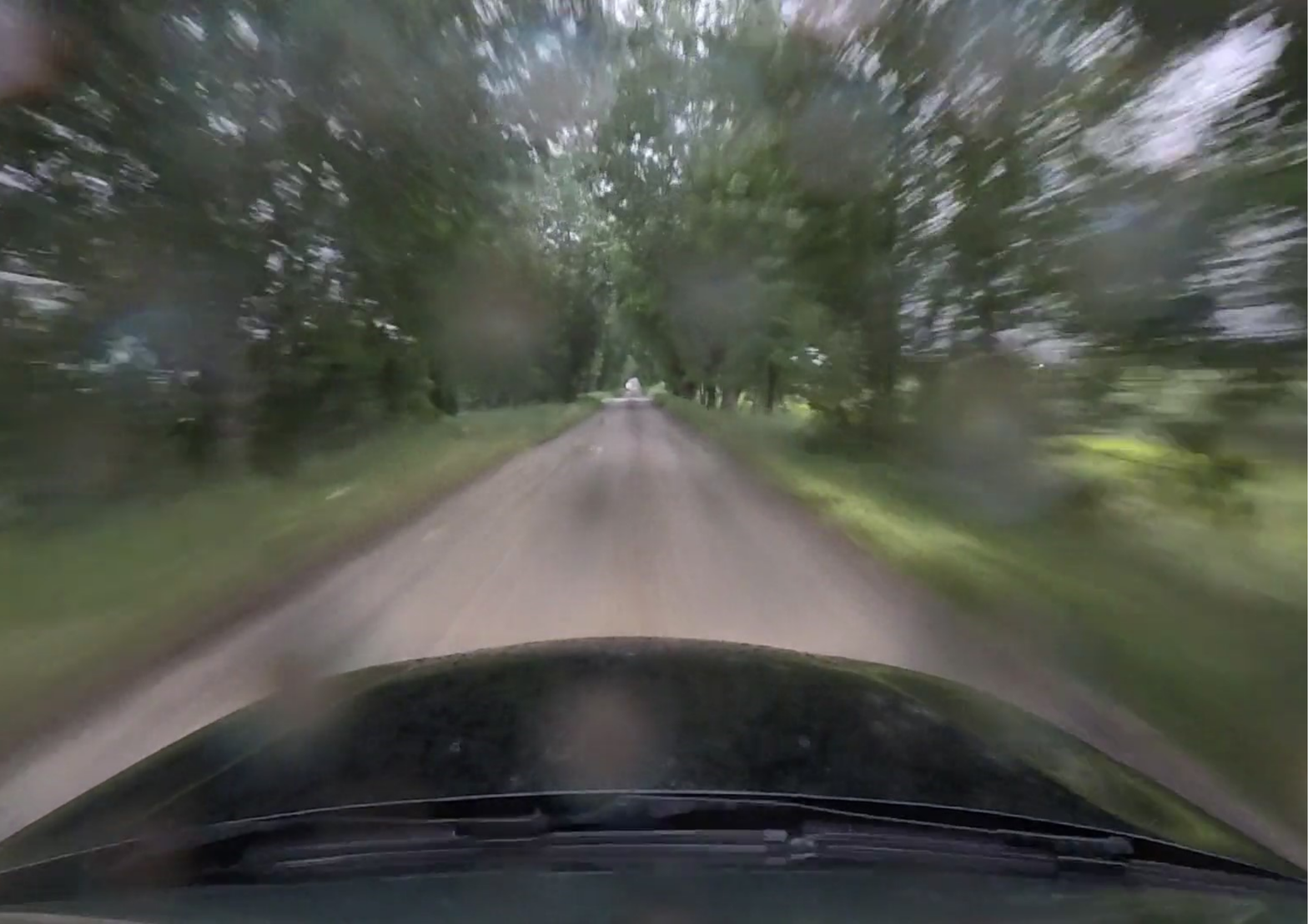}
  \hfill
  \includegraphics[width=0.32\linewidth]{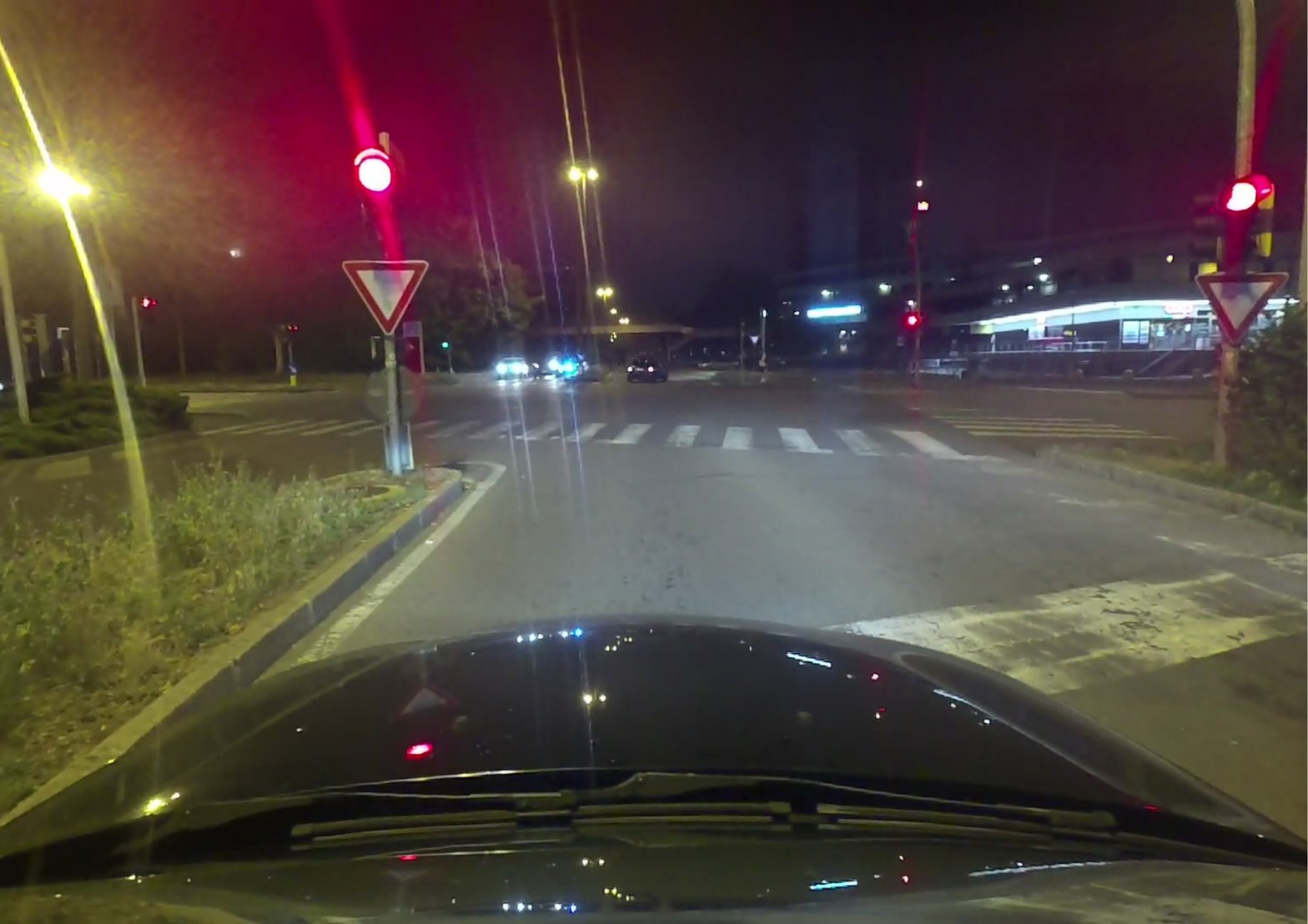}
  \caption{Example images from the DR(eye)VE dataset: (left) image from the inlier set, including sunny roads, followed by two outlier samples of rain (middle) and night driving (right).}
  \label{fig:dreyeve-examples}
\end{figure}

The fourth dataset has data from DR(eye)VE \cite{dreyeve2016}, which consists of images from real life driving. The data is comprised of 74 videos with sequences of five minutes each with annotation of drivers' gaze fixations. The dataset contains meta data describing daylight condition, weather, and driving situation. Using this meta data we partition the dataset into inlier/outlier with daylight and night-driving/rainy-weather as in- and outliers respectively, as can be seen in Fig.~\ref{fig:dreyeve-examples}.

\begin{table}
\centering
\caption{Description of the data used for each of the test cases. For the Pro-SiVIC cases, the training set consist of highway images, and outliers are represented by foggy or urban scenarios. For DR(eye)VE the training set consist of daylight driving without rain, and the outlier set is represented by night driving or rain.
%\lars{ Why different training samples across the Dreyve data?}} \jens{Different use of MetaData to exclude overlapping  data. I add this to the description
}
\begin{tabular}{lllll}
\textbf{Case} & \textbf{\begin{tabular}[c]{@{}l@{}}Training\\ samples\end{tabular}} & \textbf{\begin{tabular}[c]{@{}l@{}}Test\\ samples\end{tabular}} & \textbf{\begin{tabular}[c]{@{}l@{}}Outlier\\ samples\end{tabular}} & \textbf{\begin{tabular}[c]{@{}l@{}}Original\\ dimension\end{tabular}} \\ \hline
\rule{0pt}{3ex}MNIST/Omniglot & 60,000 & 10,000 & 10,000 & (28x28) \\
\begin{tabular}[c]{@{}l@{}}CIFAR-10/\\ Tiny ImageNet\end{tabular} & 50,000 & 10,000 & 10,000 & (32x32) \\
%Pro-SiVIC Highway & 7,629 & 788 & 787 & (752x480) \\
\begin{tabular}[c]{@{}l@{}}Pro-SiVIC Highway \\Sunny/Foggy \end{tabular} & 7,629 & 788 & 787 & (752x480) \\
%Pro-SiVIC Tunnel & 7,629 & 788 & 788 & (752x480) \\
\begin{tabular}[c]{@{}l@{}}Pro-SiVIC  Highway/\\Urban \end{tabular} & 7,629 & 788 & 488 & (752x480) \\
DR(eye)VE Night & 120,000 & 127,500 & 75,000 & (1920x1080)\\
DR(eye)VE Rain & 120,000 & 127,500 & 60,000 & (1920x1080)\\
\end{tabular}
\label{tab:data_sets}
\end{table}

\textbf{Evaluation metrics}:
In this section we present the metrics used in our framework. Our proposal is general and does not rely on any particular implementation details of the supervisors under study. We only assume that the output of the supervisor is a single number -- the \emph{anomaly score}, which measures the similarity of the input data to samples in the training dataset. In the following, we will refer to an inlier/outlier as a negative/positive data point. 

In our framework, the evaluation of a supervisor is based on four plots and seven scalar values (metrics) that show different aspects of the supervisor performance. We present them in the following and justify their individual value. 
%These metrics have been chosen based on what is standard in the field to capture the important aspects. In addition, with varying thresholds the method also reports breakpoints such as coverage at specific target levels. 

The plots consist of 1) the Receiver Operating Characteristic (ROC) curve. The curve shows how the true/false positive ratios change when the threshold of the allowed anomaly score varies. This is used to determine the discriminative capability of the supervisor with respect to positive/negative examples. In real-world applications for example, a common problem exists with skewed or imbalanced data. To better represent performance in imbalanced datasets, the precision-recall (PR) curve is more informative than the ROC-curve, and plot 2) is providing the PR curve \cite{davis-roc-vs-prc}. Although the ROC/PR curves are informative on their own, they only reveal aggregated information about the distribution of the anomaly scores across positive and negative data points. Plot 3) is hence the anomaly score distributions of the supervisor algorithm for the in- and outlier data, respectively. 
%The distribution is visualized with the test and outlier data. 
If the two distributions are clearly separable, the supervisor can easily distinguish between in- and outliers. 

So far we have only considered the performance of the supervisor as an anomaly detector and not considered its effect on the performance of the DL model under supervision. Hence, we need to incorporate the performance measure of the DL algorithm in our evaluation process. This is achieved through 4) the risk-coverage curve \cite{NIPS2017_7073}, which shows the trade-off between prediction failure rate (risk) and amount of covered data samples. Having this plot we can visualize how the performance varies as less and less data is accepted by the supervisor. 

By using the plots above, we aim at assessing a supervisor's performance from a holistic perspective. However, to compare different supervisors we need to quantify this information using comparable metrics. To this end we consider (i) the area under the ROC- (AUROC) and (ii) PR-curve (AUPRC), respectively, which represent the trade-off between true and false positives as less and less data is accepted by the supervisor. The actual shape of the ROC/PR curve is also interesting to capture, as it is preferable having its shape to be located closely to the upper left/right corner. To capture this, the third measure is (iii) true positive rate at 5\% false positive rate (TPR05) and (iv) precision at 95\% recall (P95). One desired property of the supervisor is to assign a low anomaly score to as many outliers as possible. The opposite behaviour, being over-confident that an outlier belongs to the inlier dataset, is clearly dangerous for safety-critical applications. We propose to measure this using (v) false negative rate at 95\% false positive rate (FNR95), or in other words, how many anomalies are not excluded by the supervisor when it excludes 95\% of the inlier data.

When introducing out-of-distribution samples, the performance level is reduced drastically as the output for an outlier sample is always considered erroneous. This behaviour is not captured in the metrics derived from the anomaly score distributions or the ROC/PR curves; it can be visualized, however, in the risk-coverage curve. It is clearly of interest to see how restrictive the supervisor has to be to recover the original performance on the dataset including outlier samples, or if it is not achievable with the given supervisor. We call this metric (vi) coverage breakpoint at performance level (CBPL). Finally, in similar fashion to (v), it is of interest to see if the supervisor can completely exclude outliers from the data, i.e., can we achieve full anomaly detection for some non-trivial (zero) value of coverage. We call this metric (vii) coverage breakpoint at full anomaly detection (CBFAD).

% Metrics: 
% \begin{itemize}
%     \item AUROC
%     \item AUPRC
%     \item False positive rate at \# \% True positive rate
%     \item Coverage breakpoint at \# \% Accuracy level
%     \item Coverage breakpoint at \# \% Anomaly detection  
%     \item FNR at 95\% FPR (measure of how overconfident the cage is)
% \end{itemize}

% Plots:
% \begin{itemize}
%     \item ROC - curve
%     \item Risk/Coverage 
%     \item Data distribution of anomaly score (Motivate this via reconstruction score / anomaly score from model output, etc) 
%     \item (PR - curve)
% \end{itemize}

\textbf{Models}: 
For supervisor methods that utilize information from the DL model under supervision, e.g., from bottlenecks or output from the final layers of a deep neural network, it is important to use the same model when comparing results. Some supervisors can utilize probes inside the network, which is acceptable as long as all supervisors have access to the same information. The model under supervision may change between different use cases, but must remain fixed within a use case.

\section{Example Use Cases} \label{sec:examples}
In this section, we give examples of different supervisor implementations and show how our proposed framework applies to a wide range of applications.

\subsection{Softmax Threshold for CIFAR-10/Tiny ImageNet}
In the following, we demonstrate how the information from a DNN's final layer can be used as a supervisor. The assumption is that values from the network's softmax layer will carry sufficient information as whether or not the sample is an outlier. A simple safety cage can be implemented, which considers the maximum of the predicted class distribution of the softmax layer of the network as anomaly score \cite{hendrycks2016baseline}. The hypothesis is that samples from an outlier distribution will have uncertain class results, and thus, there will be no clear class prediction. Hence we define the anomaly score as $1-\max_i(p_i)$, where $p_i$ is the softmax output for class $i$. 

\begin{figure}
\includegraphics[width=\columnwidth]{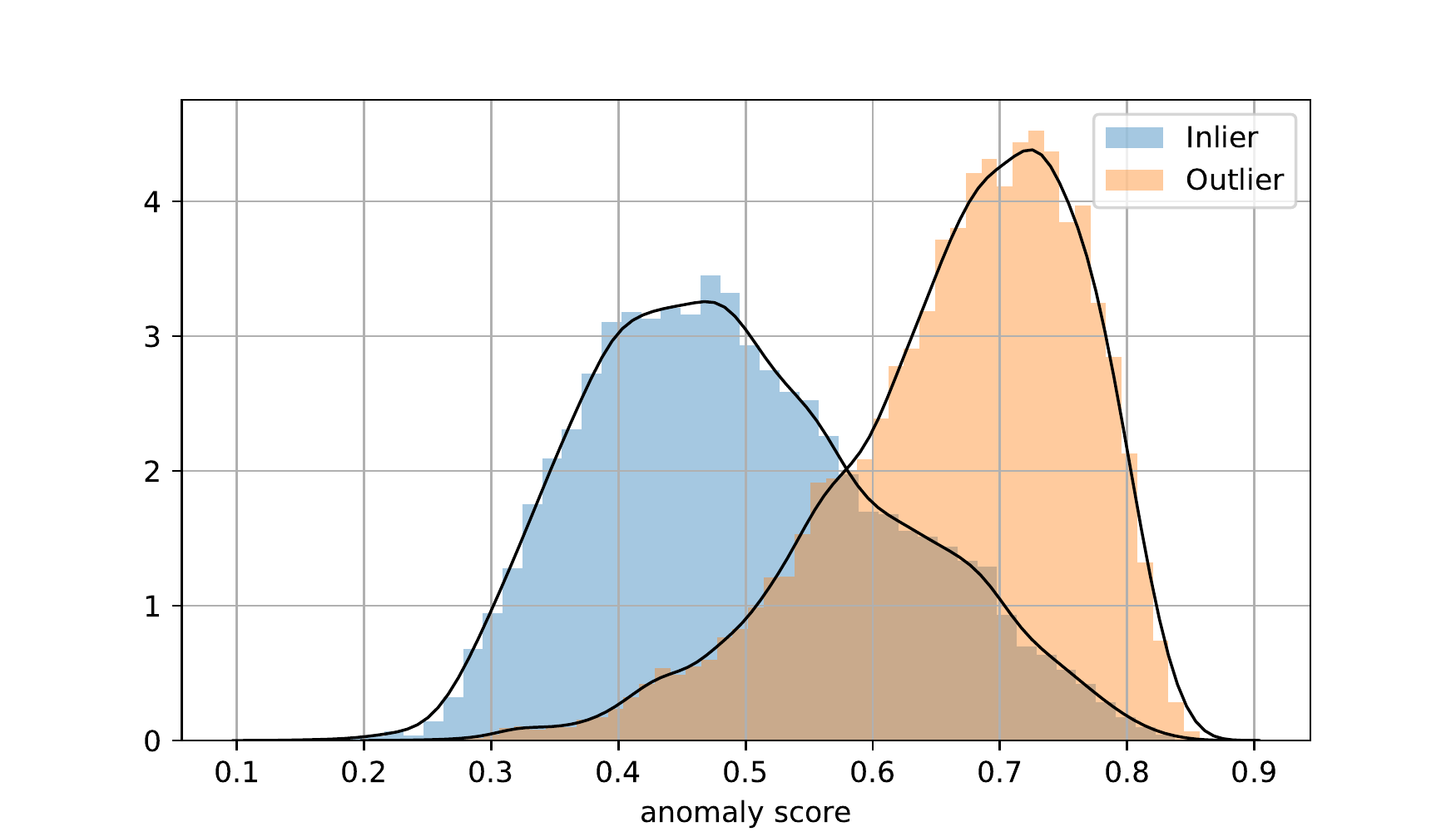}
\caption{The anomaly score with CIFAR-10 as inlier and Tiny ImageNet as outlier. The anomaly score is represented as $1-\max_i(p_i)$, where $p_i$ is the soft-max output for class $i$ from a VGG16 neural network.}
\label{fig:cifar10_anomaly_distr}
\end{figure}

\begin{figure}
\includegraphics[width=\columnwidth]{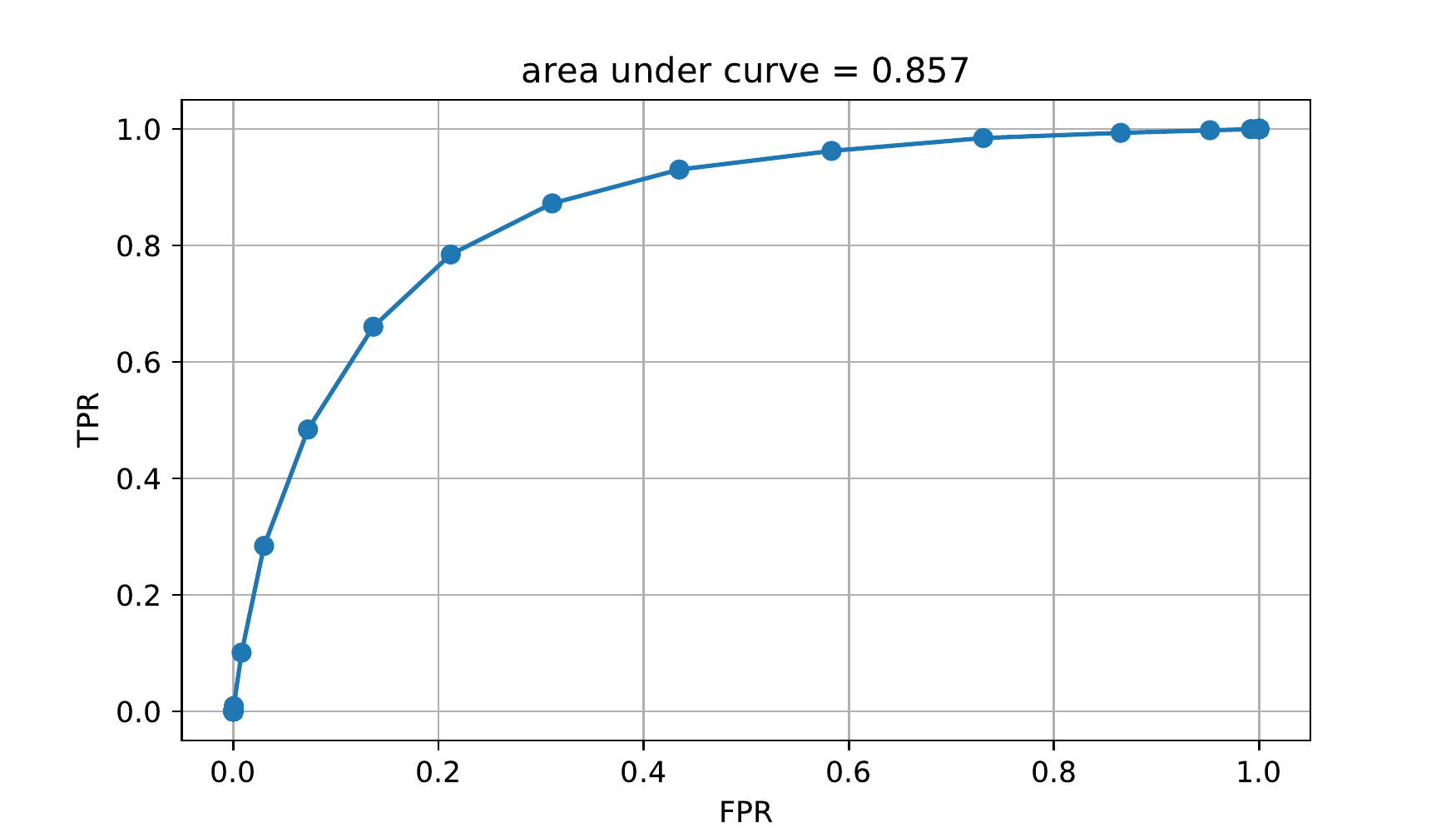}
\caption{The ROC curve corresponding to the anomaly score distributions in Fig. \ref{fig:cifar10_anomaly_distr}, with AUC $0.857$ for the CIFAR 10/Tiny ImageNet case.}
\label{fig:cifar10_roc}
\end{figure}

To evaluate this supervisor, we use the data in the CIFAR-10/Tiny ImageNet setup as described in Sec.~\ref{sec:eval}. The model under consideration is a VGG16 \cite{DBLP:journals/corr/SimonyanZ14a} network, which has been trained on CIFAR-10 to achieve $91.9\%$ accuracy. The Tiny ImageNet dataset contains 200 classes, including some that also exist in CIFAR-10, such as animals and vehicles. Due to this overlap, it is expected that anomaly score distributions for the inlier and outlier data will be overlapping as it can be seen in Fig.~\ref{fig:cifar10_anomaly_distr}. This information is also apparent in the ROC-curve in Fig.~\ref{fig:cifar10_roc}. Since the datasets are balanced, it is sufficient with the ROC-curve, and the PR-curve is excluded. From Fig.~\ref{fig:cifar10_anomaly_distr}, however, we get additional information about where to focus our efforts to improve the supervisor. Looking at the distributions we see that if we could eliminate all outliers with anomaly score less than $0.4$, we would be able to improve the situation significantly. Inspecting the data in that region could potentially also reveal why the current supervisor is failing for those images. 
%TheIt can be seen there is no clear way to distinguish between the two datasets without additional information for the supervisor, as seen in the ROC-curve in Fig. \ref{fig:cifar10_roc},  

\begin{figure}
\includegraphics[width=\columnwidth]{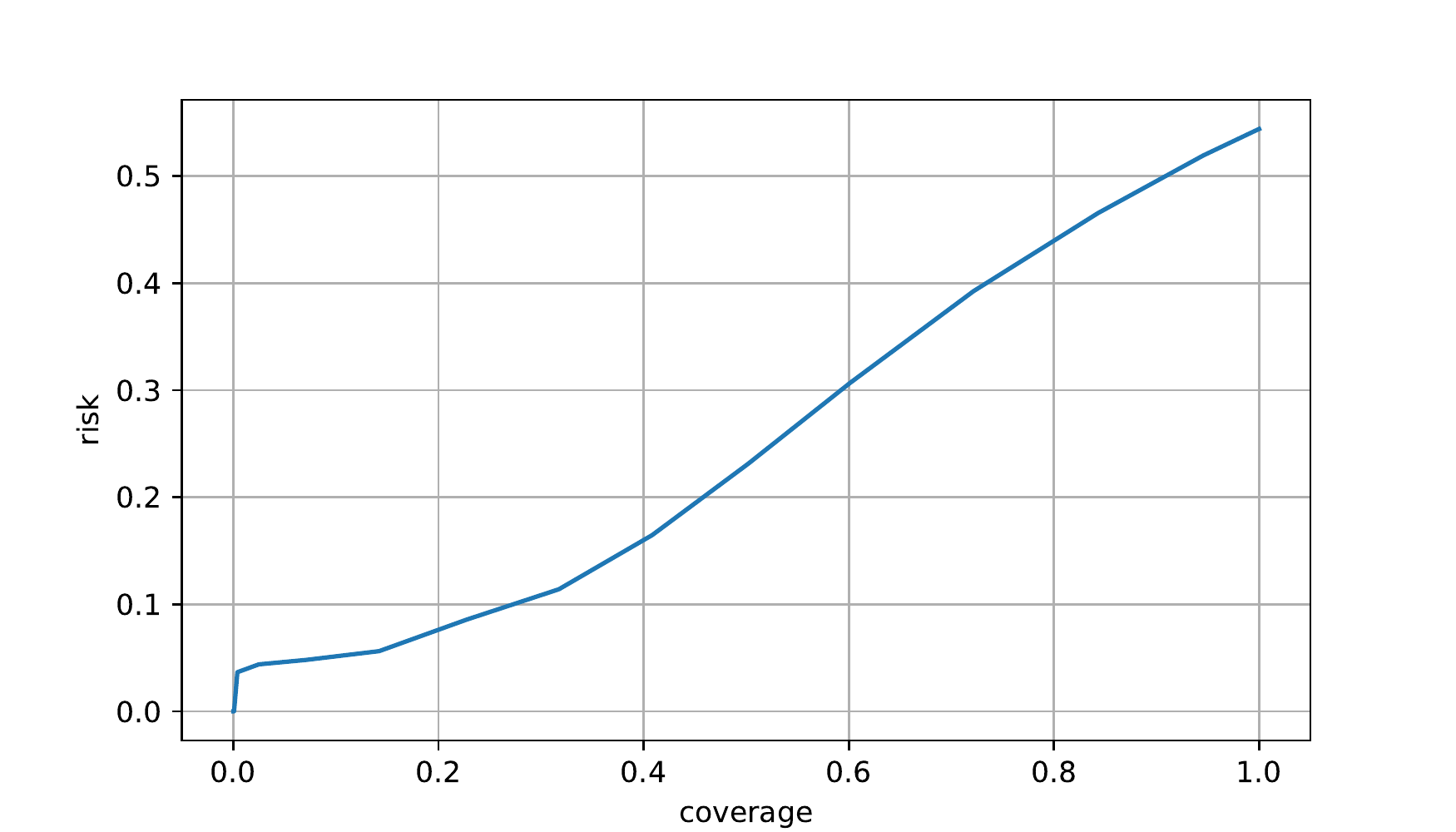}
\caption{The Risk-Coverage plot corresponding to the anomaly score distributions in Fig. \ref{fig:cifar10_anomaly_distr}}
\label{fig:cifar10_risk_coverage}
\end{figure}

For this example, the risk in the risk-coverage plot is defined as 1-accuracy thus showing the risk of mis-classifying an input sample. As seen in Fig.~\ref{fig:cifar10_risk_coverage}, the risk is monotonically decreasing with decreasing coverage, which is expected since all outliers are mis-classified per definition. %Therefor the supervisor becomes more confident in the prediction it does, at the cost of classifying less samples. 
%In similar fashion, the remaining outlier samples in the test case can be seen in Fig. \ref{fig:cifar10_outlier_coverage}, where almost $80\%$ of the outliers are removed at $50\% coverage$.  
One interesting thing to note is that there is non-zero risk for zero coverage, which means that the model is making erroneous classifications for data points with the lowest anomaly score. Hence the maximum softmax prediction does not reflect the true correctness likelihood meaning that the network is badly conditioned \cite{DBLP:conf/icml/GuoPSW17}. This indicates that using this particular anomaly score and supervisor for this network might not be optimal. This is very useful information that we would have missed if just looking at AUROC as the performance metric for example. The values of all evaluation metrics are given in Tab.~\ref{tab:evaluation_metrics}.
%\begin{figure}
%\includegraphics[width=0.48\textwidth]{CIFAR10-TinyImageNet/Outlier_vs_coverage.pdf}
%\caption{The corresponding Outlier removals vs coverage \lars{we don't have this plot in our evaluation list. Should we remove it?}}
%\label{fig:cifar10_outlier_coverage}
%\end{figure}

\subsection{Variational Auto Encoder as Supervisor for Realistic Drive Data}
\begin{figure} 
    \centering
    \includegraphics[width=0.48\textwidth]{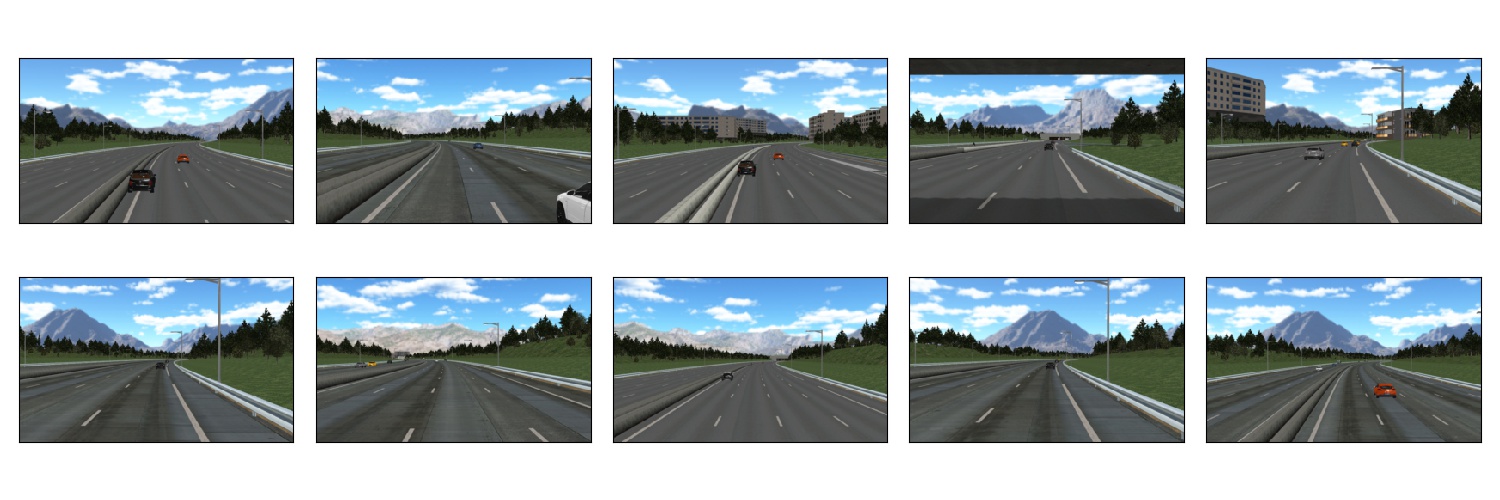}
    \caption{Image samples from the inlier set for the realistic drive scenarios generated from ESI Pro-SiVIC\textsuperscript{TM}. The inlier data represent highway driving in sunny weather conditions.}
    \label{fig:procivic_inlier}
\end{figure}

\begin{figure}
    \centering
    \includegraphics[width=0.48\textwidth]{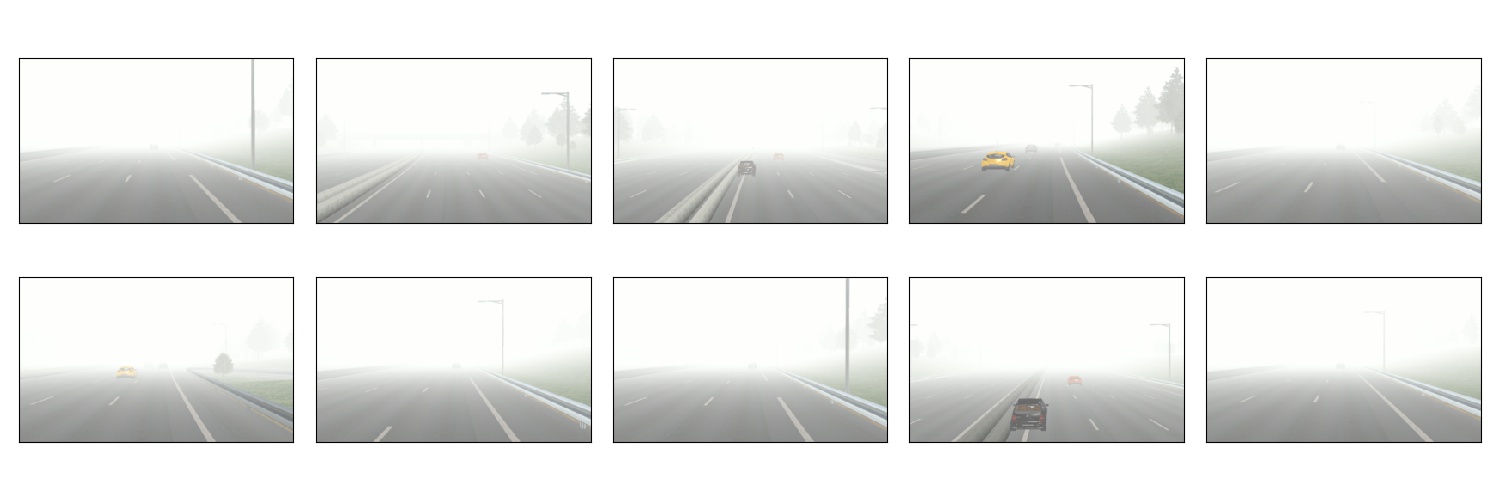}
    \caption{Image samples from the first outlier set for the realistic drive scenarios generated from ESI Pro-SiVIC\textsuperscript{TM}. Here, the outliers represent high way driving in foggy weather conditions.}
    \label{fig:procivic_outlier_fog}
\end{figure}

\begin{figure}
    \centering
    \includegraphics[width=0.48\textwidth]{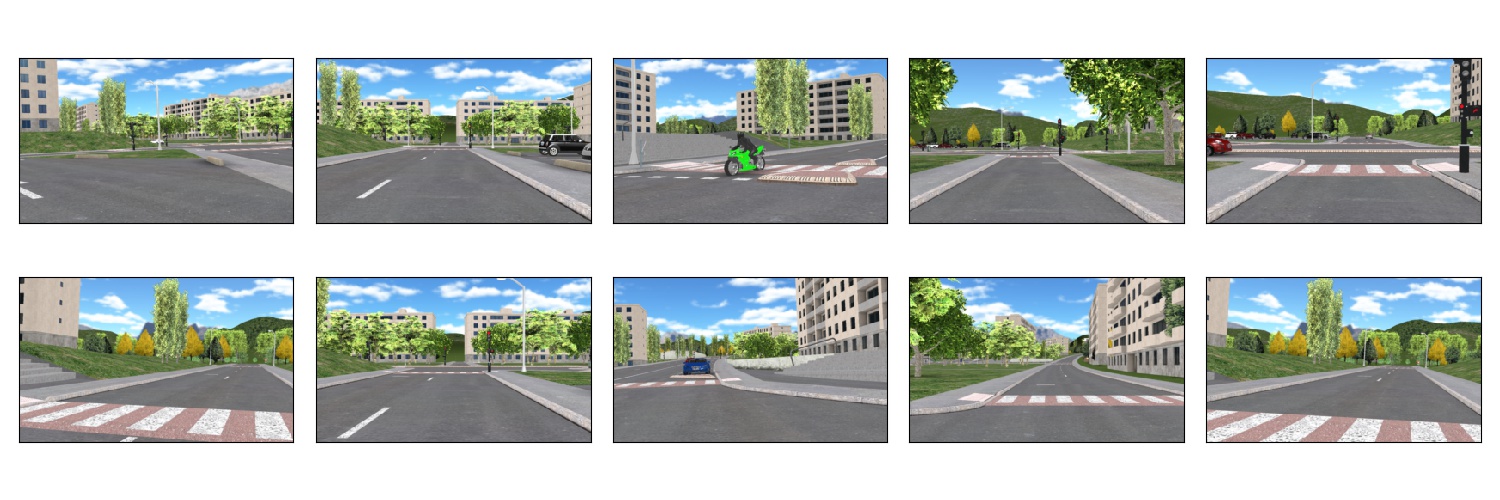}
    \caption{Image samples from the second outlier set for the realistic drive scenarios generated from ESI Pro-SiVIC\textsuperscript{TM}. Here, the outliers represent city driving in urban scenarios.}
    \label{fig:procivic_outlier_urban}
\end{figure}

In this section, a Variational Auto Encoder (VAE) is used as a safety cage for detecting anomalies in realistic driving scenarios. The data is described in Sec.~\ref{sec:eval} with all images down-scaled to $320\times 192\times 3$. Samples of the inlier images can be seen in Fig.~\ref{fig:procivic_inlier} and is represented by highway scenarios in sunny weather. The inlier set contains 8,413 training and 787 test images. To evaluate the supervisor we use two sets of anomalies. The first set contains the same images as in the test set, but with fog overlaid as depicted in Fig.~\ref{fig:procivic_outlier_fog}.
The second set of anomalies contain 488 images from typical urban scenarios which can be seen in Fig.~\ref{fig:procivic_outlier_urban}.

\begin{figure}
    \centering
    \includegraphics[width=\columnwidth]{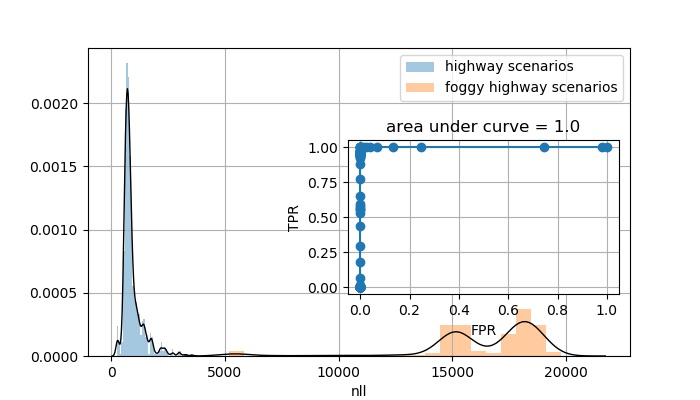}
    \caption{The distribution of the anomaly score for the inlier (high way) and outlier (fog) data. The inset shows the corresponding ROC curve with AUC = 1.} 
    \label{fig:nnl_foggyScenarios}
\end{figure}

This use case represents the scenario, where the supervisor has access only to the input data to the perception layer of e.g., an autonomous vehicle decision-and-control pipeline. Thus, we have no access to the softmax output from an image classifier as was used for anomaly score in the example before. Instead, we use a VAE as anomaly detector \cite{An2015VariationalAB}. After training, a VAE can be used both as a generative model to create new images that are representative of the training data, and also to give a lower bound of the log likelihood \cite{DBLP:journals/corr/KingmaW13} of a given image under the model. Here, we use this lower bound as an anomaly score (the negative of the log likelihood, NLL) for each input image and test the supervisor with the metrics presented in Sec.~\ref{sec:eval}. Note that since we have no access to the output of any detection or classification algorithm, we can not gauge how the supervisor would effect e.g., accuracy; hence, there is no need to consider the metrics related to risk and coverage. In this case, the supervisor can be tested using the anomaly score distribution, ROC curve and corresponding metrics only. 
In Fig.~\ref{fig:nnl_foggyScenarios} we show the distributions of the highway and foggy highway scenarios, respectively, with the inset showing the ROC curve. For this type of anomaly the supervisor shows a performance of AUROC/AUPR = 1 and thus produces perfect output. 

\begin{figure}
    \centering
    \includegraphics[width=\columnwidth]{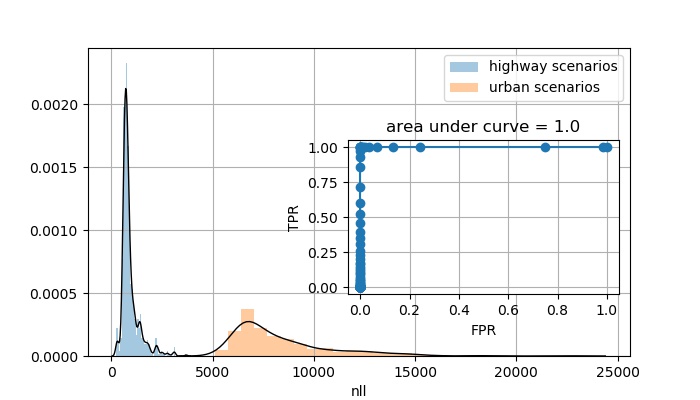}
    \caption{The distribution of the anomaly score for the inlier (high way) and outlier (urban) data. The inset shows the corresponding ROC curve with AUROC = 1. Comparing the separation of in- and outlier distributions with that in Fig.~\ref{fig:nnl_foggyScenarios}, it is clear that the urban scenarios represent a more difficult type of anomaly for this safety cage compared with the foggy scenarios.}
    \label{fig:nnl_urbanScenarios}
\end{figure}

We subject the same trained VAE as was used to detect foggy images with the urban scenarios as anomalies. In Fig.~\ref{fig:nnl_urbanScenarios} we show the distributions of the in- and outlier data, again with the inset showing the  ROC curve. Note that the same supervisor shows a performance of AUROC/AUPR = 1 for this type of anomaly as well. However, from the distribution plots it is apparent that the urban scenarios represent a more difficult set of outliers than the foggy weather condition, which is information that is lost in the aggregated data represented by the ROC curve. This is a clear illustration of the usefulness of having more than one single measure of performance.
The values of all evaluation metrics is given in Tab.~\ref{tab:evaluation_metrics}.

%%%% Table for Results

\begin{table*}[]
\centering
\caption{Summary of the evaluation metrics for the example use cases in Sec.~\ref{sec:examples}. Note that there is no performance baseline for the VAE supervisor and hence CPBL becomes N/A in this case.}
\begin{tabular}{ll|lllllll}
\multicolumn{2}{c|}{\textbf{Test setup}} & \multicolumn{7}{c}{\textbf{Metrics}} \\
\multicolumn{1}{c}{\textbf{Supervisor}} & \multicolumn{1}{c|}{\textbf{Case}} & \textbf{AUROC} & \textbf{AUPRC} & \textbf{TPR05} & \textbf{P95} & \textbf{FNR95} & \textbf{CBPL} & \textbf{CBFAD}  \\ \hline
Softmax threshold & CIFAR10 / Tiny ImageNet & 0.857 & 0.836 & 0.284 & 0.623 & 0.0022 & 0.2264 & 0 \\
Variational Autoencoder & Pro-SiVIC Highway Sunny/Foggy & 1 & 1 & 1 & 1 & 0 & N/A & 0.5 \\
Variational Autoencoder & Pro-SiVIC Highway/Urban       & 1 & 1 & 1 & 1 & 0 & N/A & 0.62
\end{tabular}
\label{tab:evaluation_metrics}
\end{table*}

%%%%%%%%%%%%%%%%%%%%%%%%%%%%%%%%%%%%%%%%%%%%%%%%%%%%%%%%%%%%%%%%%%%%%%%%%%%%%%%%%%%%%%
\section{Discussion} \label{sec:disc}
It is understood that explainability is lacking for DL-enabled systems. When exposed to out-of-distribution samples, the behaviour can be unpredictable, which requires safe argumentation of DL before deployment. Our framework defines how a comparison between supervisors can be used as one input for an argumentation of a safe development. Our hypothesis is that information from the training set, as well as characteristics from the NN can be used to limit the number of erroneous predictions that will occur if the environment undergoes changes that makes it different from the training time. 

The test cases proposed in this paper consist of a mix with public well-known datasets, which frequently appear in DL literature and hence, should be a valid target for DL supervisor research. These cases can be extended further to cover additional datasets, which are relevant for a particular application. As an example, this paper covers cases geared towards self-driving vehicular applications with the addition of realistic driving data and thereby demonstrating the transferability of the approach to datasets with different structure and complexity.  

We are not covering how to design supervisors in this paper. They can be designed to use either network parameters, training data, or both, as long as bias from the outlier set is not introduced in the supervisors. Whether or not network training details such as overfitting, batch normalization, drop out, etc.~are affecting the ability to supervise the model is also not covered. This could possibly yield additional insight of how to establish safe argumentation for DL deployment.

Another aspect to consider is to base the supervisor on deep learning itself. It introduces, however, an interesting catch-22 logic: If the DL component is supervised by another DL component, has anything been gained from a safety perspective? On the other hand, introducing redundancy, e.g., two independently implemented components providing similar functionality, is a common technique in safety engineering~\cite{leaphart2005survey}. As for now, we suggest that all options should be explored further to better understand the limitations of supervisors for DL-enabled systems.   

Finally, all of the proposed datasets and supervisors work as feed forward networks i.e., no recursiveness or memory is considered. The images in the realistic driving datasets consist of time series data, which would allow usage of historical information to increase performance of the supervisor.

%%%%%%%%%%%%%%%%%%%%%%%%%%%%%%%%%%%%%%%%%%%%%%%%%%%%%%%%%%%%%%%%%%%%%%%%%%%%%%%%%%%%%%

\section{Conclusion and future work} \label{sec:conc}
In this paper, an evaluation framework for deep neural network supervisors is described. Our framework defines the evaluation metrics and proposes test cases to enable comparison between supervisors. In addition, we show how to apply two supervisors to two example use cases exhibiting different complexity.

The outcome of our work opens several directions for future work. First, our approach will pave the way for systematic comparisons of different supervisors. Our plan is to pursue such comparative work next, initially with supervisors mentioned in Sec.~\ref{sec:rw}. From this comparison, the characteristics that define a good supervisor can be derived and used to synthesize an improved version. 

Second, the evaluation approach is modular in the sense that additional supervisors, as well as other types of DL tests can be added or integrated. Furthermore, a selection of test cases can be used for task specific solutions since some cases can be irrelevant for others. In future work, we plan to develop guidelines for how to best choose between the test cases provided in our framework. 

Third, we believe that supervisor evaluation using a state-of-the-art simulator is a feasible target for metamorphic testing~\cite{chen1998metamorphic}. By using a simulator, we can generate data adhering to metamorphic relations such as different weather conditions, alternative camera positions or small image transformations. For some metamorphic relations we know that a supervisor should yield the same results, e.g., changing the color of surrounding traffic should not be considered as novel situation. On the other hand, depending on how the normal operating conditions are defined, adding heavy fog or changing from paved roads to gravel roads could be used to test if the supervisor triggers. We believe our approach to supervisor testing is scalable, and more controllable than using GANs to generate images, for which metamorphic relations hold as implemented in DeepRoad~\cite{zhang2018deeproad}.

Finally, in one of our research projects\footnote{https://www.viktoria.se/projects/smile-ii}, we are developing a safety case relying on a supervisor, also known as a safety cage architecture~\cite{heckemann_safe_2011,adler_safety_2016,varshney_engineering_2016}. This supervisor will be evaluated with the approach described in this paper.

\section*{Acknowledgments}
This work was carried out within the SMILE II project financed by Vinnova, FFI, Fordonsstrategisk forskning och innovation under the grant number: 2017-03066, and partially supported by the Wallenberg AI, Autonomous Systems and Software Program (WASP) funded by Knut and Alice Wallenberg Foundation.

\bibliographystyle{IEEEtran}
\bibliography{smile}

\end{document}